\pdfoutput=1

\documentclass[11pt]{article}

\usepackage[]{EMNLP2022}

\usepackage{times}

\usepackage{latexsym}
\usepackage{amsfonts,amssymb}
\usepackage{amsmath}
\usepackage{graphicx}
\usepackage{enumitem}
\usepackage{multirow}
\usepackage[english]{babel}
\newtheorem{theorem}{Theorem}

\usepackage{booktabs}
\usepackage{caption}
\usepackage{float} 
\usepackage{subcaption}

\usepackage[T1]{fontenc}

\usepackage[utf8]{inputenc}

\usepackage{microtype}

\usepackage{inconsolata}

\title{Unsupervised Non-transferable Text Classification}

\author{
 Guangtao Zeng \and Wei Lu
 \\
StatNLP Research Group\\
Singapore University of Technology and Design\\
 \texttt{guangtao\_zeng@mymail.sutd.edu.sg, luwei@sutd.edu.sg}
}

\begin{document}
\maketitle
\begin{abstract}
Training a good deep learning model requires substantial data and computing resources, which makes the resulting neural model a valuable intellectual property.
To prevent the neural network from being undesirably exploited, non-transferable learning has been proposed to reduce the model generalization ability in specific target domains.
However, existing approaches require labeled data for the target domain which can be difficult to obtain.
Furthermore, they do not have the mechanism to still recover the model's ability to access the target domain.
In this paper, we propose a novel unsupervised non-transferable learning method for the text classification task that does not require annotated target domain data.
We further introduce a {\em secret key} component in our approach for recovering access to the target domain, where we design both an {\em explicit} and an {\em implicit} method for doing so.
Extensive experiments demonstrate the effectiveness of our approach.
\end{abstract}

\section{Introduction}
Deep learning has achieved remarkable success over the past decade and is active in various fields, including computer vision, natural language processing (NLP), and data mining.
Although neural models can perform well in most tasks, they require a huge amount of data and a high computation cost to train, making the trained model a valuable intellectual property.
As a result, it is essential for us to prevent neural models from being used without authorization.
In the last few years, many methods have been proposed to safeguard  deep neural networks and they can be roughly divided into two types: {\em watermarking}~\cite{DBLP:conf/uss/AdiBCPK18}, and {\em secure authorization}~\cite{DBLP:journals/corr/abs-2008-05966}.
In the watermarking approaches, the owners can verify the ownership of the neural model based on a unique watermark.
However, due to the catastrophic forgetting problem \cite{kemker2018measuring}, the watermark-based neural models \cite{kuribayashi2020deepwatermark, song2017machine} {\color{black} are known to be vulnerable to  certain} malicious attacks~\cite{wang2019attacks}, {\color{black} which may lead to the loss of their watermarks}.
On the other hand, in the secure authorization approaches, the owners of the neural network want to ensure that users can only access the model with authorization.
Recently, \citet{wang2022nontransferable} proposed a new perspective with non-transferable learning (NTL) to protect the model from illegal access to unauthorized data.
The method trains the model to have good performance only in the authorized domain while performing badly in the unauthorized domain.
However, such an approach has some limitations: 
{\color{black} 1) their work relies on a significant amount of labeled data from the target domain, while such labels are usually not easy to acquire,}
2) the access to the unauthorized domain can no longer be regained, if required, after {\color{black} the model is learned}.

\begin{figure}[t!]\hspace{-0.2cm}
    \centering
    \includegraphics[width=0.4\textwidth]{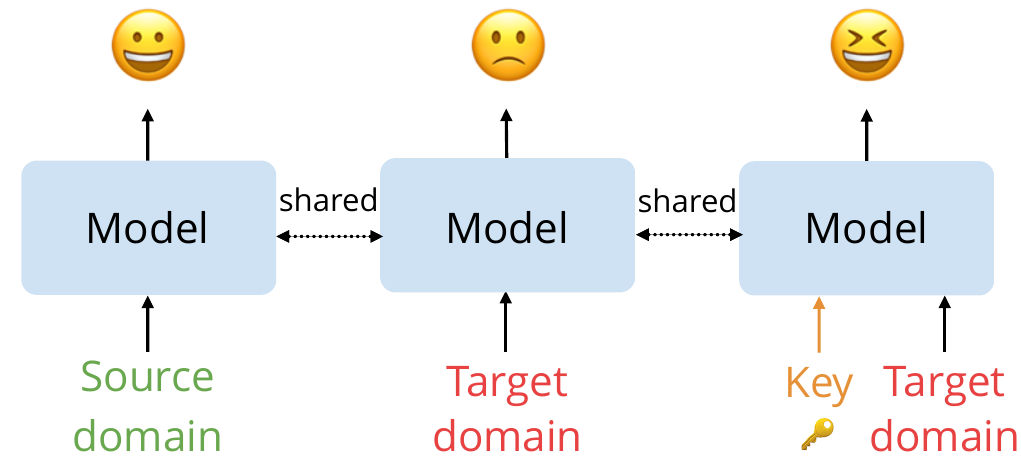}
    \caption{Overview of our unsupervised non-transferable learning method with secret keys.}
    \label{fig:intro}
\end{figure}

{\color{black} In this work, we propose our new NTL method named \underline{U}nsupervised \underline{N}on-\underline{T}ransferable \underline{L}earning (UNTL) for the text classification tasks.}
As Figure~\ref{fig:intro} shows, our model can perform well in the source domain while performing badly in the target domain. In addition, we propose {\color{black} secret key modules}, which can help recover the ability of the model in the target domain.
Our contributions include:

 \begin{itemize}
  \item
We propose a novel unsupervised non-transferable learning approach for text classification tasks. 
{\color{black} Different from existing approaches, our model can still perform well without the need for the label information in the target domain.}

  \item
We introduce two different methods, {\color{black} namely {\em Prompt-based Secret Key} and {\em Adapter-based Secret Key}, that} allow us to recover the ability of the model to perform classification on the target domain.
  \item
{\color{black} Extensive experiments show that our proposed models can perform well in the source domain but badly in the target domain. Moreover, access to the target domain can still be regained using the secret key.}

\end{itemize}

{\color{black}
To the best of our knowledge, our work is the first approach for learning under the unsupervised non-transferable learning setup, which also comes with the ability to recover access to the target domain.\footnote{Our code and data are released at \url{https://github.com/ChaosCodes/UNTL}.}}

\section{Related Work}

In this section, we briefly survey ideas that are related to our work from two fields: domain adaptation and intellectual property protection. Furthermore, we discuss some limitations in the existing methods which we will tackle with our approach.

In domain adaptation, given a source domain and a target domain with unlabeled data or a few labeled data, the goal is to improve the performance in the target task using the knowledge from the source domain.
\citet{ghifary2014domain}, \citet{tzeng2014deep}, and \citet{zhu2020deep} applied a Maximum Mean Discrepancy regularization method~\cite{gretton2012kernel} to maximize the invariance information between different domains.
\citet{ganin2016domain} and \citet{schoenauer-sebag2018multidomain} tried to match the feature space distributions of the two domains with adversarial learning.
In contrast to the methods above, \citet{wang2022nontransferable} analyzed domain adaptation in a different way and proposed non-transferable learning (NTL) to prevent the knowledge transfer from the source to the target domain by enlarging the discrepancy between the representations in different domains.

In intellectual property protection, due to the significant value and its vulnerability against malicious attacks of learned deep neural networks, it is crucial to propose intellectual property protection methods to defend the owners of the deep neural networks (DNNs) from any loss.
Recently, two different approaches to safeguard DNNs have been proposed: watermarking~\cite{DBLP:conf/uss/AdiBCPK18} and secure authorization~\cite{DBLP:journals/corr/abs-2008-05966}.
In the watermarking approaches, researchers designed a digital watermark that can be embedded into data such as video, images, and so on.
With the detection of the unique watermark, we could verify the ownership of the copyright of the data.
Based on these ideas, \citet{song2017machine} and \citet{kuribayashi2020deepwatermark} embedded the digital watermarks into the parameters of the neural networks.
\citet{zhang2020model} and \citet{wu2020watermarking} proposed a framework to generate images with an invisible but extractable watermark.
However, they are vulnerable to some active attack algorithms~\cite{wang2019attacks, DBLP:conf/asiaccs/Chen0BDJLS21} which first detect the watermark and then rewrite or remove it.
On the other hand, the secure authorization approach seeks to train a model that generates inaccurate results without authorization.
\citet{DBLP:journals/corr/abs-2008-05966} proposed a key-based framework that ensures correct model functioning only with the correct secret key.
In addition, \citet{wang2022nontransferable} were inspired by domain generalization and proposed non-transferable learning (NTL), which achieves secure authorization by reducing the model's generalization ability in the specified unauthorized domain.

Although the NTL model can effectively prevent access to the unauthorized domain, it requires target labels during training, which may not always be easy to obtain.
Furthermore, there is no mechanism to recover access to the unauthorized domain when needed.
In this paper, we present a new NTL model and show that our model can still have good performance even in the absence of the target labels which are, however, indispensable in the work of \citet{wang2022nontransferable}.
Besides, we extend it to a secret key-based version. With our method, authorized users can still access the target domain with the provided keys.
\section{Approach}
In this section, we first introduce our proposed {U}nsupervised {N}on-{T}ransferable {L}earning  (UNTL) approach in Sec.~\ref{sec:UNTL}, followed by a discussion on its practical limitation -- it lacks the ability to regain the access to the target domain.
Next, we discuss {\color{black}our secret key-based methods} in Sec.~\ref{sec:key_methods} to address this limitation. \looseness=-1

\subsection{UNTL Text Classification}\label{sec:UNTL}

\paragraph{Problem Description}\label{description}
First of all, we present our definition of the unsupervised non-transferable learning task without labeled data from the target domain.
\textcolor{black}{Following~\citet{DBLP:journals/corr/abs-2010-03978}, we consider that a domain consists of three parts: input space $X$, label space $Y$, and the joint probability distribution $p(X, Y)$.}
Given a source domain $\mathcal{D}_s=\{\boldsymbol{x}_i,\boldsymbol{y}_i\}_{i=1}^{N_s}$ and a target domain $\mathcal{D}_t=\{\boldsymbol{x}_j\}_{j=1}^{N_t}$ with unlabeled samples, where $\boldsymbol{y}_i \in \mathbb{R}^C$ is a one-hot vector indicating the label of $\boldsymbol{x}_i$, $C$ is the number of classes, and $N_s, N_t$ refer to the number of examples in the source and target domain respectively.
The goal of our UNTL method is to prevent the knowledge transfer from the source to the target domain, i.e.,  to train the model so that it performs well on the source domain data but poorly on the target domain data, without the requirement of accessing the label information of the target data. 

\paragraph{Text Classification} In our work, we use a BERT-based model~\cite{devlin2018bert} $\psi$ as our feature extractor for the input sentence and consider the final hidden state $h$ of the token \texttt{[CLS]} as the feature representation, where we denote $h$ as $\psi\left(\boldsymbol{x}\right)$.
A simple feed-forward network ${\fontfamily{lmr}{\text{FFN}}}\left(\cdot \right)$ will be added on top of BERT as a classifier to predict the label.
The formal loss function can be:
\begin{equation}
\mathcal{L}_\textrm{CE} = \mathbb{E}_{\left(\boldsymbol{x}, \boldsymbol{y}\right)\sim\mathcal{D}_s}[ {\fontfamily{lmr}{\text{CE}}}\left({\fontfamily{lmr}{\text{FFN}}}\left(\psi\left(\boldsymbol{x} \right) \right), \boldsymbol{y}\right)]
\end{equation}
where $\mathcal{D}_s$ is the source domain dataset and ${\fontfamily{lmr}{\text{CE}}}$ indicates the cross entropy function.

\paragraph{Maximum Mean Discrepancy}
To enlarge the distance between the representations of the source and target domains, we follow \citet{wang2022nontransferable} and use Maximum Mean Discrepancy~\cite{gretton2012kernel} (MMD) to achieve this goal. MMD is a kernel two-sample test and can be used as a metric to determine whether two data distributions $p$ and $q$ are similar.
MMD defines the metric function as follows:
\begin{equation}\label{eq:mmd1}
d_{{p,q}} = ||\mathbb{E}_{\boldsymbol{x}\sim p}[\psi\left(\boldsymbol{x}\right)] - \mathbb{E}_{\boldsymbol{x}'\sim q}[\psi\left(\boldsymbol{x}'\right)]||^2_{\mathcal{H}_k}
\end{equation}
{\color{black} where $\mathcal{H}_k$ is the reproducing kernel Hilbert space (RKHS) with a kernel $k$, whose operation is $k\left(\boldsymbol{z}, \boldsymbol{z}'\right) = e^{-||\boldsymbol{z}-\boldsymbol{z}'||^2}$ and function $\psi$ maps the sentence input into  RKHS.
The smaller the distance $d_{p,q}$, the more similar the two distributions $p$ and $q$.}\looseness=-1

In our work, we use MMD to increase the distance between the feature representations of the source and the target domain, forcing the feature extractor $\psi$ to extract domain-dependent representations rather than maximizing the inter-domain invariance.
To prevent the high MMD from dominating the entire loss, we follow \citet{wang2022nontransferable} and set an upper bound for it.
Therefore, based on Equation~\ref{eq:mmd1}, our MMD loss can be formulated as:
\begin{equation}
\mathcal{L}_{\textrm{MMD}}\left(\mathcal{S,T}\right) = -\min\left(c, {d}_{\mathcal{S,T}}\right)
\end{equation}
where $c$ is the upper bound for MMD, and {\color{black}$\mathcal{S,T}$ are data distributions of the source and target domains respectively.}
With this loss, we only maximize ${d}_{\mathcal{S,T}}$ when it is smaller than the upper bound $c$.

\paragraph{Domain Classifier}
Despite being able to enlarge the gap between the source and target domains to some extent, 
the MMD loss lacks the explicit ability to clearly draw the boundary between the representations of different domains, especially when the knowledge between domains is similar.
{\color{black}
Therefore, we hypothesize that using MMD alone may not be sufficient to yield optimal empirical performance.}
To mitigate this issue, we draw inspiration from the Domain-Adversarial Neural Networks~\cite{ganin2016domain} and propose an additional domain classifier added on top of the feature extractor.
This classifier is trained to predict the domain with the feature representations.
We employ a cross-entropy loss to train the domain classifier.
By optimizing this loss, the representations of different domains are encouraged to be more distinct. 
Specifically, we use 0 to indicate the source domain and 1 to indicate the target domain. We can formulate the domain classification (DC) loss as:
{\color{black}
\begin{equation}
\begin{aligned}
\mathcal{L}_{\textrm{DC}}\left(\mathcal{S}, \mathcal{T}\right) =   \mathbb{E}_{\boldsymbol{x}^S\sim \mathcal{S}}[{\fontfamily{lmr}{\text{CE}}}\left({\fontfamily{lmr}{\text{FFN}}_{dc}}\left(\psi\left(\boldsymbol{x}^S\right)\right), 0\right)]   \\
    +\ \mathbb{E}_{\boldsymbol{x}^T\sim\mathcal{T}}[{\fontfamily{lmr}{\text{CE}}}({\fontfamily{lmr}{\text{FFN}}_{dc}}\left(\psi\left(\boldsymbol{x}^T\right)\right), 1)]
\end{aligned}
\end{equation}
where ${\fontfamily{lmr}{\text{FFN}}_{dc}}$ is the domain classifier}.
With this DC loss as a regularization term, the boundary of feature representation between the source and the target can be clearer, facilitating better non-transferable learning.

\paragraph{Objective Function}
In this task, our goal is to train a model that can perform well on the source domain while performing badly on the target domain. 
To achieve this goal, we propose a loss function for unsupervised non-transferable learning, which contains three terms.
The first term is the cross-entropy loss $\mathcal{L}_{\textrm{CE}}$ for text classification to integrate knowledge about the downstream task into the model.
The second term is the domain classification loss $\mathcal{L}_{\textrm{DC}}$ and the third is MMD loss $\mathcal{L}_{\textrm{MMD}}$.
The latter two terms jointly contribute to enlarging the gap between the representations of the source and target domains to prevent knowledge transfer. 
Finally, we can get our overall loss which is written as:
\begin{equation}\label{eq:total}
\mathcal{L}_{\textrm{UNTL}} = \mathcal{L}_{\textrm{CE}} +\beta\cdot \mathcal{L}_{\textrm{DC}}+ \lambda\cdot\mathcal{L}_{\textrm{MMD}}
\end{equation}
where $\beta$ and $\lambda$ are the scaling hyperparameters.
\paragraph{Theoretical Analysis}
{\color{black} Different from \cite{wang2022nontransferable} where they use information bottleneck theory \cite{DBLP:journals/corr/physics-0004057} to show the feasibility of non-transferable learning, we turn to a more general theory of domain adaptation}~\cite{ben2006analysis, wang2018theoretical}.
Here, we present an analysis of the effectiveness of the unsupervised setting based on this theory.

\begin{theorem}~\cite{ben2010theory} Let $\mathcal{H}$ be a hypothesis space (of a particular VC dimension), for any $h \in \mathcal{H}$. Given a source domain $\mathcal{D_S}$ and a target domain $\mathcal{D_T}$:
\begin{equation} \label{eq:theorem1}
    \epsilon_{\mathcal{T}}\left(h\right) \le \epsilon_{\mathcal{S}}\left(h\right) + \frac{1}{2} d_{\mathcal{H}\Delta\mathcal{H}}\left(\mathcal{D}_{\mathcal{S}},\mathcal{D}_{\mathcal{T}}\right) + C
\end{equation}
where $\epsilon_{\mathcal{S}}\left(h\right)$ and $\epsilon_{\mathcal{T}}\left(h\right)$ are the expected source and target errors respectively,  $C = \min\limits_{h' \in \mathcal{H}} \left(\epsilon_{\mathcal{S}}\left(h'\right) + \epsilon_{\mathcal{T}}\left(h'\right)\right)$, which can be viewed as a constant and
$d_{\mathcal{H}\Delta\mathcal{H}}$ is a divergence\footnote{$d_{\mathcal{H}\Delta\mathcal{H}}$ is a symmetric difference hypothesis space for a hypothesis space $\mathcal{H}$. See \citet{ben2010theory} for more details.} that measures the maximal discrepancy between two distributions under a fixed hypothesis class.
\end{theorem}

{\color{black}
During our training process, we minimize the source error while maximizing the divergence (with the MMD and DC losses).
Comparing with a baseline transfer model without the MMD and DC losses,
we hypothesize that our method may yield a comparable source error, while leading to a significantly larger divergence term.
We believe this may lead to a significant increase in the target error, as the changes of the above terms would effectively lead to a much looser upper bound for the target error as shown in Equation \ref{eq:theorem1}.
Such a looser upper bound may lead to a significant increase in target error, which can effectively prevent the knowledge from being transferred into the target domain.
We will verify our hypothesis in the experiments later.\footnote{In fact, through our experiments later, we found that on average there was a 1\% increase in source error for our approach as compared to the baseline. However, there was a significant increase ($\times 10$) in the divergence term as approximated by the MMD loss, which leads to effective non-transfer learning (where we achieve good source domain performance and bad target domain performance).}}

\subsection{Learning with Secret Keys} \label{sec:key_methods}

 With our UNTL method, we could ensure that the model performs well in the source domain whilst degrading its performance in the target domain.
 However, it is inconvenient if the performance in the target domain can no longer be restored after training.
 This can be illustrated with an example: suppose that we are running an application that supports two kinds of users, regular users, and members.
 Suppose further that the regular users are only authorized to query the model for a limited set of data, while the members have no limits on their access to the model.
 Using our UNTL approach discussed above, we can limit the access of the regular users by denoting the authorized and non-authorized portions of the data as the source and target domains respectively.
 Then we train the model that performs well on the source domain, but poorly on the target domain.
 However, as the members have no limits to their access, they would require a separate model to be trained that performs well on both domains, thus doubling the computational and storage costs required for the application.

To solve this issue, we extend our approach to include a secret key, $K$, that can be used to recover access to the target domain even after non-transferable learning.
Without the key, the model is encouraged to perform well on the source domain while degrading the accuracy in the target domain.
However, upon supplying the secret key, the model's performance on the target domain will be restored.
Following the example above, this allows a single neural network to be used for all the users, whilst providing privileged access to the members through the provision of the secret key.
 Based on our UNTL, in this section, we present our innovative secret key-based unsupervised non-transferable learning method. 
 The method can not only keep the normal users away from the membership area but also provide members with a specific key that allows them to access the membership area within a single model.
 
 Our intuition is to design a secret key that can revive the restricted target domain in our UNTL model.
 We call the method Secret Key-based Non-Transferable Learning, which has two variants$\colon$
 1) {\em Prompt-based Secret Key method}, where we add a discrete prompt as a prefix to the input sentence that serves as an explicit secret key to restore access to the target domain,
 and 2) {\em Adapter-based Secret Key method},  where a trained adapter module is added to the model to transform the target embeddings into the source-like ones in an implicit manner.

\begin{figure}[t!]
    \centering
    \includegraphics[width=0.45\textwidth]{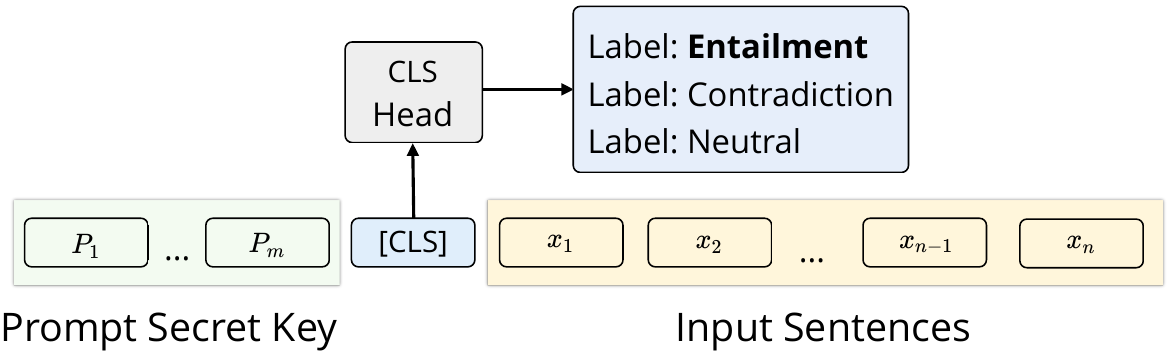}
    \caption{Prompt-based Secret Key Methods}
    \label{fig:prompt}
\end{figure}

\paragraph{Prompt-based Secret Key}

Recently, prompt-based learning \cite{DBLP:conf/eacl/SchickS21, DBLP:conf/emnlp/LesterAC21} has achieved state-of-the-art results in many tasks.
Inspired by these prompt-based methods, we consider a prompt as our secret key, which users can use to access the target domain.
As shown in Figure~\ref{fig:prompt}, we first assign a randomly chosen prompt $P=\{P_1, ..., P_m \}$ as the secret key, where $P_i$ is the $i$-th token in the prompt sentence and $m$ is the length of the prompt.
Given an input sentence $\boldsymbol{x}$ with length $n$, we concatenate the prompt with the input $\boldsymbol{x}$ to construct an authorized input sentence.
In addition, similar to inference without the prompt key, we continue using the hidden representation at the position of \texttt{[CLS]} as the input of the task classifier and get the predicted label. 

With the introduction of the prompt-based key, we believe that there are 3 different distributions in this task: {\em source} domain, {\em target} domain, and {\em target+prompt} domain.
In the prompt-based secret key model, after prepending the specific prompt to the target input, the model can recover the ability to perform well in the target domain.
Therefore, we try to train the feature extractor to close the distance between the target+prompt domain and the source domain while enlarging the distance between the source and the target domain without the key.
To achieve this, we propose a new MMD loss:
\begin{equation}
\mathcal{L'}_{\textrm{MMD}}\left(\mathcal{P,S,T}\right) = \alpha \cdot d_{\mathcal{P, S}} - \min\left(c, d_{\mathcal{S,T}}\right)
\end{equation}
{\color{black} where $\mathcal{P}$ denotes the data distribution of} the target+prompt domain, $\alpha$ is the scaling hyperparameter and $c$ is the upper bound for MMD.

In this way, we can transfer the knowledge from the source domain to the target+prompt domain but not to the original target domain.
Therefore, we can extend Equation~\ref{eq:total} to get the objective function for the prompt-based secret key UNTL method:
\begin{equation} \label{eq:prompt}
\begin{aligned}
\mathcal{L}_{\textrm{prompt}} =& \mathcal{L}_\textrm{CE} + \beta \cdot \mathcal{L}_\textrm{DC}\left(\left[\mathcal{P}, \mathcal{S}\right], \mathcal{T}\right) + \\ & \lambda \cdot \mathcal{L'}_{\textrm{MMD}}\left(\mathcal{P,S,T}\right)
\end{aligned}
\end{equation}
where $\left[\mathcal{P}, \mathcal{S}\right]$ indicates {\color{black} the data distribution of the combined domain of} source and target+prompt.

\paragraph{Adapter-based Secret Key}
\begin{figure}[t!]
    \centering
    \includegraphics[width=0.3\textwidth]{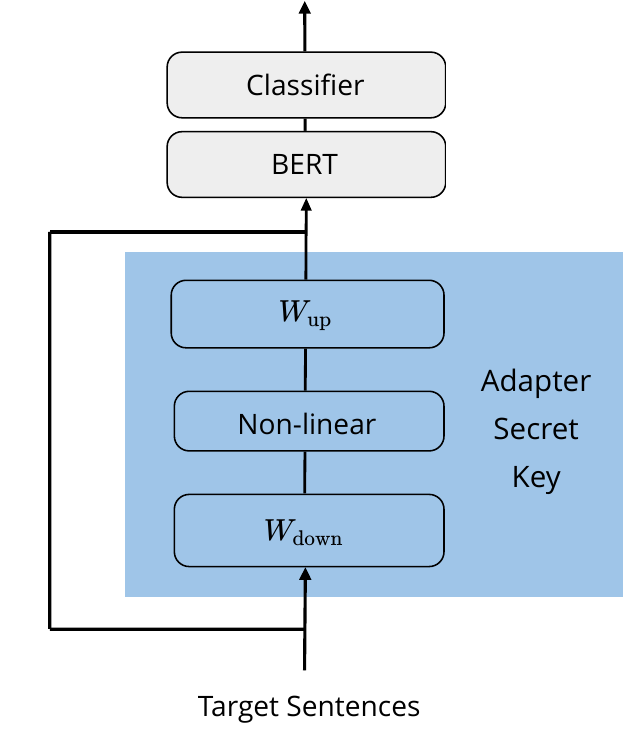}
    \caption{Adapter-based Secret Key Methods}
    \label{fig:adapter}
\end{figure}
Besides explicitly prepending the input sentences with the discrete prompt as the secret key, we could also consider adding an input adapter~\cite{houlsby2019parameter, an2022input} as the secret key.
In our UNTL model, input sentences in different domains will lead to distinct performance.
Intuitively, we train the input adapter to eliminate the target-like feature and convert the target embeddings into source-like embeddings.
{\color{black}
Given an embedding representation, we assume it has two components: the semantic component which is essential for text classification, and the domain component which is irrelevant to the task. We train the adapter to convert the domain component of the embedding from the target domain to the source domain while maintaining the semantic component.}

{\color{black}The adapter architecture is shown in Figure~\ref{fig:adapter}. In the figure, embeddings of the target sentences are first projected into a lower dimensional space $\mathbb{R}^m$  with $W_{\textrm{down}}$ before passing through a ReLU nonlinear function, and then projected back to the space $\mathbb{R}^d$ with $W_{\textrm{up}}$, where $m$ is significantly less than $d$ (in addition, there is a skip connection as shown in Figure~\ref{fig:adapter}).}
With the input adapter module, the target embeddings will be transformed into source-like ones.
From this adapter-based network, we can get source-like embedding data in the {\em target+adapter} domain which can be denoted as $\mathcal{D}_{\mathcal{A}}$.
Similar to the prompt-based secret key method, we inject the knowledge from the source domain into the target+adapter domain by closing the distance $d_{\mathcal{A}, \mathcal{S}}$ between their representations.

However, directly using the UNTL loss above could not make sure that the adapter can maintain the task-dependent information.
{\color{black}Therefore, we construct a dataset $\{{\fontfamily{lmr}{\text{adapter}}}\left(\boldsymbol{x}_i\right), \boldsymbol{y}_i\}^{N_s}_{i=1}$ with the source domain data, where ${\fontfamily{lmr}{\text{adapter}}}\left(\boldsymbol{x}\right)$ indicates the representation of $\boldsymbol{x}$ converted by the adapter module.
Then we train the model with an additional cross-entropy loss to guarantee that the embeddings converted by the adapter can contain sufficient signals about the classification task:}
\begin{equation}
\begin{aligned}
&\mathcal{L}_{
\textrm{CE}_{\textrm{adapter}}
} = \\ 
&\mathbb{E}_{\left(\boldsymbol{x}, \boldsymbol{y}\right)\sim\mathcal{D}_s}\left[{\fontfamily{lmr}{\text{CE}}}\left({\fontfamily{lmr}{\text{FFN}}}\left( \psi\left(
{\fontfamily{lmr}{\text{adapter}}}
(\boldsymbol{x}\right)\right)\right), \boldsymbol{y})\right]
\end{aligned}
\end{equation}

The overall objective function for training the adapter-based secret key model is:
\begin{equation} \label{eq:adapter_model}
\begin{aligned}
\mathcal{L}_{\textrm{adapter}} =& \mathcal{L}_\textrm{CE} + \mathcal{L}_{\textrm{CE}_{\textrm{adapter}}} + \beta\cdot\mathcal{L}_\textrm{DC}\left(\left[\mathcal{A}, \mathcal{S}\right], \mathcal{T}\right)\\
& + \lambda\cdot\mathcal{L'}_{\textrm{MMD}}\left(\mathcal{A,S,T}\right)
\end{aligned}
\end{equation}
where $\left[\mathcal{A}, \mathcal{S}\right]$ indicates the {\color{black} data distribution} of the combined domain of source and target+adapter.
\begin{table}[t!]
\normalsize
\setlength\tabcolsep{2pt}
\centering
\scalebox{0.8}{
\begin{tabular}{l ccccc}
\toprule
{Datasets} &\multicolumn{1}{c}{\textsc{SL}}&\multicolumn{1}{c}{\textsc{TE}}&\multicolumn{1}{c}{\textsc{GO}}&\multicolumn{1}{c}{\textsc{TR}}&\multicolumn{1}{c}{\textsc{FI}}\\
\midrule
\#Train& 68,716 & 74,087 & 68,755 & 68,755 & 68,753\\
\#Valid& {\color{white}0}8,590 & {\color{white}0}9,261 & {\color{white}0}8,595 & {\color{white}0}8,595 & {\color{white}0}8,595\\
\#Test& {\color{white}0}1,955 & {\color{white}0}1,966 & {\color{white}0}1,945 & {\color{white}0}1,976 & {\color{white}0}1,973\\
\bottomrule
\end{tabular}
}
\caption{Dataset statistics}
\label{tab:multinli}
\end{table}
\section{Experiments}

In this section, we first conduct experiments to verify the
effectiveness of our UNTL method and then show that our proposed secret key-based UNTL approach could recover access to the target unauthorized domain when the secret key is supplied in the form of a discrete prompt or an additional adapter module.
We use MultiNLI~\cite{williams2017broad} as our benchmark dataset, {\color{black} which is for a 3-class classification task with balanced labels.}
We begin with some training details of all experiments and then discuss the results in different settings.

\subsection{Experimental Setup}
Our models are implemented in PyTorch~\cite{DBLP:conf/nips/PaszkeGMLBCKLGA19} and all experiments are conducted with NVIDIA Quadro RTX 8000 GPUs and we run three times with different seeds. We use the preprocessed MultiNLI dataset from Huggingface~\cite{lhoest2021datasets}.
Based on the genre information, we divide the MultiNLI dataset into 5 parts, namely, slate (\textsc{SL}), telephone (\textsc{TE}), government (\textsc{GO}), travel (\textsc{TR}), and fiction (\textsc{FI}) as different domains.
As Huggingface only has a training set and a validation set for MultiNLI, we split the training dataset into 8:1 as the training set and the evaluation set, and consider the validation set as the test set in our experiments.
The dataset statistics can be found in Table~\ref{tab:multinli}.
As for the model architecture, we use the pretrained language model BERT~\cite{devlin2018bert} as our feature extractor and a randomly initialized one-layer feed-forward network as the classifier.
We use the Adam~\cite{kingma2014adam} optimizer with $\beta_1$ = 0.9, $\beta_2$ = 0.999.
More details can be found in Appendix~\ref{sec:appendix}.
\begin{table}[t!]
\small
\setlength\tabcolsep{3pt}
\centering
\scalebox{0.95}{
\begin{tabular}{l  c c c c c}
\toprule
 Src\textbackslash Tgt & \textsc{SL}  & \textsc{TE} & \textsc{GO} & \textsc{TR} & \textsc{FI} \\
\midrule
 \multicolumn{1}{c}{\textsc{SL}} & $73.8_{\pm 0.6}$ & $73.9_{\pm 0.4}$  & $80.1_{\pm 0.2}$  &$76.6_{\pm 0.3}$  & $76.8_{\pm 0.6}$  \\
 \multicolumn{1}{c}{\textsc{TE}} & $72.1_{\pm 0.4}$ & $77.5_{\pm 0.6}$  & $77.1_{\pm 0.2}$  &$74.0_{\pm 0.3}$  & $75.3_{\pm 0.9}$  \\
 \multicolumn{1}{c}{\textsc{GO}} & $71.7_{\pm 0.3}$ & $72.0_{\pm 0.3}$  & $81.0_{\pm 0.2}$  &$75.9_{\pm 0.8}$  & $74.1_{\pm 0.5}$  \\
  \multicolumn{1}{c}{\textsc{TR}} & $71.6_{\pm 0.5}$ & $71.6_{\pm 0.1}$  & $80.4_{\pm 0.3}$  &$79.3_{\pm 0.3}$  & $73.9_{\pm 0.4}$  \\
   \multicolumn{1}{c}{\textsc{FI}} & $73.4_{\pm 0.6}$ & $74.2_{\pm 0.2}$  & $79.2_{\pm 0.4}$  &$75.4_{\pm 0.4}$  & $80.1_{\pm 0.6}$  \\
\bottomrule
\end{tabular}
}
\caption{Performance of classification task in each domain when training on the source domain only}
\label{tab:sup}
\end{table}

\begin{table}[t!]
\small
\setlength\tabcolsep{3pt}
\centering
\scalebox{0.95}{
\begin{tabular}{l c c c c c}
\toprule
 Src\textbackslash Tgt & \textsc{SL}  & \textsc{TE} & \textsc{GO} & \textsc{TR} & \textsc{FI} \\
\midrule
 \multicolumn{1}{c}{\textsc{SL}} & $71.5_{\pm 1.4}$ & $34.2_{\pm 1.1}$  & $40.0_{\pm 0.7}$  &$34.1_{\pm 1.7}$  & $36.6_{\pm 1.5}$  \\
 \multicolumn{1}{c}{\textsc{TE}} & $33.7_{\pm 1.3}$ & $76.8_{\pm 0.5}$  & $35.4_{\pm 1.8}$  &$34.7_{\pm 1.0}$  & $32.6_{\pm 0.5}$  \\

 \multicolumn{1}{c}{\textsc{GO}}& $38.0_{\pm 0.8}$ & $34.4_{\pm 1.1}$  & $81.1_{\pm 0.8}$  &$34.1_{\pm 0.1}$  & $34.9_{\pm 1.4}$  \\

 \multicolumn{1}{c}{\textsc{TR}} & $36.9_{\pm 3.8}$ & $33.4_{\pm 0.1}$  & $34.6_{\pm 1.9}$  &$78.9_{\pm 0.8}$  & $33.6_{\pm 0.1}$  \\
 \multicolumn{1}{c}{\textsc{FI}} & $39.1_{\pm 1.8}$ & $34.6_{\pm 1.0}$  & $36.0_{\pm 1.8}$  &$34.8_{\pm 0.8}$  & $78.7_{\pm 1.3}$ \\
\bottomrule
\end{tabular}
}
\caption{Performance over the target domain for our unsupervised non-transferable learning}
\label{tab:ntl}
\end{table}

\begin{table}[t!]
\small
\setlength\tabcolsep{3pt}
\centering
\scalebox{0.95}{
\begin{tabular}{l c c c c c}
\toprule
 Src\textbackslash Tgt & \textsc{SL}  & \textsc{TE} & \textsc{GO} & \textsc{TR} & \textsc{FI} \\
\midrule
 \multicolumn{1}{c}{\textsc{SL}} & 
$72.7_{\pm 1.1}$ &  $33.0_{ \pm 0.7 }$  &  $38.1_{ \pm 1.5 }$  &  $36.2_{ \pm 1.3 }$  &  $38.0_{ \pm 0.0 }$ \\ 
 \multicolumn{1}{c}{\textsc{TE}} & 
 $34.7_{ \pm 0.7 }$  & $76.7_{\pm 0.7}$ &  $34.7_{ \pm 2.7 }$  &  $34.1_{ \pm 1.7 }$  &  $34.5_{ \pm 1.1 }$ \\

 \multicolumn{1}{c}{\textsc{GO}}& 
 $37.4_{ \pm 1.1 }$  &  $32.3_{ \pm 0.8 }$  & $80.8_{\pm 0.4}$ &  $33.8_{ \pm 1.6 }$  &  $33.3_{ \pm 0.5 }$ \\ 
 \multicolumn{1}{c}{\textsc{TR}} & 
 $36.2_{ \pm 1.6 }$  &  $33.4_{ \pm 0.0 }$  &  $35.1_{ \pm 2.8 }$  & $79.2_{\pm 0.6}$ &  $34.3_{ \pm 1.1 }$ \\
 
 \multicolumn{1}{c}{\textsc{FI}} & 
 $38.1_{ \pm 1.0 }$  &  $34.0_{ \pm 1.8 }$  &  $33.6_{ \pm 2.8 }$  &  $34.0_{ \pm 1.9 }$  & $79.0_{\pm 0.7}$\\ 
 
\bottomrule
\end{tabular}
}
\caption{Performance over the target domain for non-transferable learning with the label information in the target domain.
The results are on par with the UNTL results, which suggests that source labels alone are sufficient for the UNTL task.}
\label{tab:addition_ntl}
\end{table}

\begin{table*}[t!]
\small
\setlength\tabcolsep{3pt}
\centering
\scalebox{0.92}{
\begin{tabular}{l c  c c c c  }
\toprule
Source\textbackslash Target & \textsc{SL}  & \textsc{TE} & \textsc{GO} & \textsc{TR} & \textsc{FI} \\
\midrule
\multicolumn{1}{c}{\textsc{SL}} & $72.8_{\pm 0.9} \Rightarrow 68.6_{\pm 1.3}$ & 
$35.3_{\pm 1.0} \Rightarrow 68.9_{\pm 0.3}$  & $38.2_{\pm 1.6} \Rightarrow 74.7_{\pm 0.7}$  &$36.9_{\pm 1.7} \Rightarrow 73.1_{\pm 1.0}$  & $39.2_{\pm 1.4} \Rightarrow 70.2_{\pm 1.4}$  \\
\multicolumn{1}{c}{\textsc{TE}} & $33.1_{\pm 1.3} \Rightarrow 65.1_{\pm 1.7}$ &
$76.4_{\pm 1.1} \Rightarrow 71.1_{\pm 1.2}$  
& $33.5_{\pm 2.3} \Rightarrow 75.4_{\pm 0.3}$  &$34.0_{\pm 1.0} \Rightarrow 71.6_{\pm 0.4}$  & $34.2_{\pm 1.6} \Rightarrow 70.8_{\pm 0.4}$  \\

\multicolumn{1}{c}{\textsc{GO}} & $38.8_{\pm 1.5} \Rightarrow 63.5_{\pm 0.6}$ & $32.9_{\pm 0.7} \Rightarrow 66.5_{\pm 0.6}$ & 
$80.8_{\pm 0.9} \Rightarrow 76.8_{\pm 1.5}$  
&$34.5_{\pm 1.4} \Rightarrow 72.0_{\pm 0.7}$  & $35.6_{\pm 0.7} \Rightarrow 66.2_{\pm 1.2}$  \\

\multicolumn{1}{c}{\textsc{TR}} & $37.2_{\pm 0.8} \Rightarrow 62.6_{\pm 0.8}$ & $35.4_{\pm 0.1} \Rightarrow 66.2_{\pm 1.4}$  & $34.1_{\pm 2.5} \Rightarrow 77.7_{\pm 0.3}$  & $78.8_{\pm 0.7} \Rightarrow 73.8_{\pm 1.3}$  & $36.2_{\pm 0.7} \Rightarrow 66.4_{\pm 0.5}$  \\

\multicolumn{1}{c}{\textsc{FI}} & $41.5_{\pm 0.6} \Rightarrow 67.2_{\pm 0.6}$ & $34.7_{\pm 1.0} \Rightarrow 68.8_{\pm 0.7}$  & $37.3_{\pm 0.1} \Rightarrow 76.3_{\pm 0.4}$  & $36.4_{\pm 0.2} \Rightarrow 71.7_{\pm 0.8}$  & 
$78.8_{\pm 0.7} \Rightarrow 76.2_{\pm 0.7}$ \\
\bottomrule
\end{tabular}
}
\caption{Performance of prompt-based secret key NTL model. The left of the right arrow shows the accuracy (\%) of the model when the prompt-based key was not attached to the input, and the right is the precision with the key.}
\label{tab:prompt}
\end{table*}

\begin{table*}[t!]
\small
\setlength\tabcolsep{3pt}
\centering
\scalebox{0.92}{
\begin{tabular}{l c  c c c c}
\toprule
Source\textbackslash Target & \textsc{SL}  & \textsc{TE} & \textsc{GO} & \textsc{TR} & \textsc{FI} \\
\midrule
\multicolumn{1}{c}{\textsc{SL}} & $72.8_{\pm 0.4} \Rightarrow 73.5_{\pm 0.5}$ & $35.0_{\pm 1.0} \Rightarrow 73.5_{\pm 0.6}$  & $37.1_{\pm 2.7} \Rightarrow 79.0_{\pm 0.2}$  &$37.4_{\pm 2.6} \Rightarrow 74.8_{\pm 0.6}$  & $42.7_{\pm 2.4} \Rightarrow 76.8_{\pm 0.4}$  \\

\multicolumn{1}{c}{\textsc{TE}} & $33.2_{\pm 1.5} \Rightarrow 71.5_{\pm 0.5}$ & 
$77.0_{\pm 0.8} \Rightarrow 76.7_{\pm 0.7}$  
& $34.0_{\pm 1.8} \Rightarrow 77.8_{\pm 0.5}$  &$34.7_{\pm 1.1} \Rightarrow 74.3_{\pm 0.8}$  & $34.2_{\pm 1.6} \Rightarrow 75.4_{\pm 0.3}$  \\
\multicolumn{1}{c}{\textsc{GO}} & $41.0_{\pm 0.7} \Rightarrow 70.2_{\pm 0.7}$ & $33.9_{\pm 1.5} \Rightarrow 71.4_{\pm 0.6}$ & 
$80.1_{\pm 1.2} \Rightarrow 80.2_{\pm 1.1}$  
&$35.2_{\pm 1.2} \Rightarrow 74.4_{\pm 0.3}$  & $36.7_{\pm 1.6} \Rightarrow 73.6_{\pm 1.2}$  \\

\multicolumn{1}{c}{\textsc{TR}} & $37.9_{\pm 1.6} \Rightarrow 69.9_{\pm 0.8}$ & $34.1_{\pm 1.1} \Rightarrow 71.9_{\pm 1.0}$  & $34.4_{\pm 2.5} \Rightarrow 79.2_{\pm 0.3}$  & $79.6_{\pm 0.8} \Rightarrow 79.8_{\pm 0.6}$  &
$34.4_{\pm 1.9} \Rightarrow 73.8_{\pm 0.3}$  \\

\multicolumn{1}{c}{\textsc{FI}} & $41.2_{\pm 3.0} \Rightarrow 71.8_{\pm 0.4}$ & $34.1_{\pm 1.8} \Rightarrow 74.3_{\pm 0.3}$  & $35.9_{\pm 2.7} \Rightarrow 78.7_{\pm 0.3}$  & $35.5_{\pm 1.9} \Rightarrow 75.0_{\pm 0.4}$  & 
$79.2_{\pm 0.7} \Rightarrow 79.5_{\pm 0.5}$ \\
\bottomrule
\end{tabular}
}
\caption{Performance of adapter-based secret key NTL model. The left of the right arrow shows the accuracy (\%) of the model when the input adapter was not applied, and the right is the precision with the adapter.}
\label{tab:adapter}
\end{table*}
\subsection{Results for UNTL}


{\color{black}
We first train a supervised classification model only on the source domain, shown in Table~\ref{tab:sup}, as a baseline.
From the baseline results, we can observe that the knowledge for one domain can be easily transferred to others.
Although only trained on the source domain, the neural network shows considerable performance on the unseen domains.
In our UNTL experiments, we traverse all possible domain pairs and Table~\ref{tab:ntl} shows that the method successfully degrades the performance in the target domain to between 32.6\% and 40.0\%, which is near random choice (33.3\%) in such 3-label classification tasks.}

We can observe that the largest performance degradation is from 80.4\% to 34.6\%, in which the source-target pair is Travel and Government.
In addition, though the target accuracy can be decreased a lot, the model can still maintain a good performance in the source domain.
The maximal average drop in the source domain is only 1\%.
The results in Table~\ref{tab:ntl} suggest that our UNTL model can successfully reduce the target performance whilst maintaining a decent accuracy in the source domain even when the source and target domains are similar and without the target labels.

\paragraph{Comparison with original NTL}
\textcolor{black}{We also compare our method with the original NTL~\cite{wang2022nontransferable} to show that the labels in the target domain are not really necessary. Table~\ref{tab:addition_ntl} shows the performance when the original NTL is applied. 
As we can see from Table~\ref{tab:ntl}, our UNTL model performs similarly to the NTL method in the source domain.
Although NTL degrades the performance slightly better in the target domain as compared to UNTL, both methods successfully reduce the accuracy on the target domain to close to random chance and the difference is negligible.
Therefore, we show empirically that labels in the target domain are not strictly necessary as our UNTL model can still succeed in preventing the knowledge transfer from the source to the target domain even without the target labels.
}

\subsection{Results for UNTL with secret keys}

\paragraph{Prompt-based Secret Key}
We continue to use all possible domain pairs in our experiments, and assign a non-task-dependent sentence `\textit{Here this a password key messages, Do not tell others.}'\footnote{\color{black} Note that this sentence that serves as a secret key is intentionally ungrammatical.} as the prompt-based key.
From Table~\ref{tab:prompt}, we could see that the performance in the target domain ranges from 32.9\% to 41.5\%.
Moreover, with the specific prompt, we can successfully access the target domain and get better performance.
We further make a comparison with the baseline in Table~\ref{tab:sup}, where non-transferable learning is not used.
Though prompt-based secret key could recover the ability, the average accuracy is 7\% worse than the baseline in the target domain.

\begin{figure*}[t!]
	\centering
	\begin{subfigure}{0.24\linewidth}
		\centering
		\includegraphics[width=1.0\linewidth]{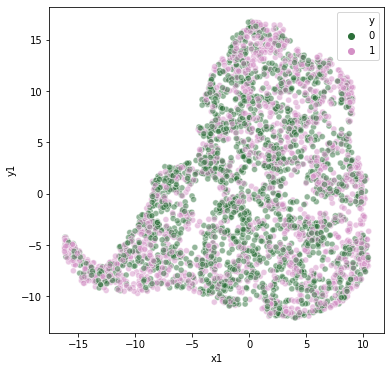}
		\caption{Transfer learning}
	\end{subfigure}
	\centering
	\begin{subfigure}{0.24\linewidth}
		\centering
		\includegraphics[width=1.0\linewidth]{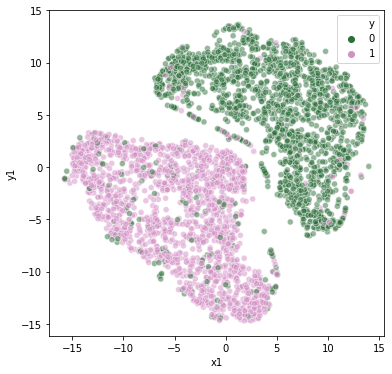}
		\caption{UNTL (w/o DC)}
	\end{subfigure}
	\centering
	\begin{subfigure}{0.24\linewidth}
		\centering
		\includegraphics[width=1.0\linewidth]{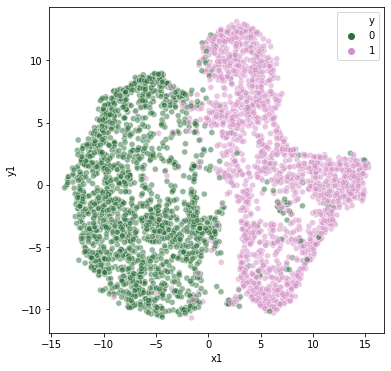}
		\caption{UNTL (w/o MMD)}
	\end{subfigure}
	\centering
	\begin{subfigure}{0.24\linewidth}
		\centering
		\includegraphics[width=1.0\linewidth]{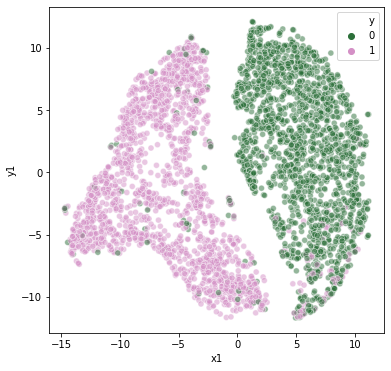}
		\caption{UNTL}
	\end{subfigure}
	\caption{Illustration of the distribution under different non-transfer methods (MMD loss and DC loss) settings}
	\label{dist}
\end{figure*}

\paragraph{Adapter-based Secret Key}
In this experiment, we apply the input adapter after the embedding layer as the secret key and train our unsupervised non-transferable learning model, and the results are shown in Table~\ref{tab:adapter}.
Under the adapter setting, the performance in the target domain can be similarly degraded to between 33.2\% and 42.7\% as with the prompt-based secret key.
Moreover, the adapter is able to restore the degraded performance in the target domain to be on par with the baseline performance in Table~\ref{tab:sup}.
{\color{black} With the additional input adapter, our method can recover the model capability in the target domain better than the prompt-based methods.}
We hypothesize the reason could be that {\color{black} the models may still struggle to distinguish the target domain from the target+prompt domain in which the instances are constructed by prepending a discrete prompt in front of the input sentences.}
The representations between them are hard to be divorced by the MMD loss and DC loss.
On the other hand, the adapter module transforms the input sentences in the continuous space and can be also jointly trained to construct a target+adapter domain that is different from the target domain.

Overall, the results demonstrate that not only we can achieve training a model that has good accuracy in the source domain while performing poorly in the target domain but also restore the performance in the target domain with an adapter. 

\paragraph{Discussion}
\textcolor{black}{Here we provide a comparison between the two types of secret keys.
We first start with the trade-offs between storage and performance. 
While the adapter-based secret key has higher storage requirements and requires an additional 99K parameters for the input adapter module, this accounts for only about 0.09\% of the BERT-base (109M parameters).
In exchange, the adapter-based secret key outperforms the prompt-based secret key by around 7\% in recovering the performance in the target domain.}

\textcolor{black}{
We also compare the performance of the model with and without the secret keys.
For the sake of simplicity, we refer to our earlier example and denote the models with and without keys as members and regular users respectively, where the members have access to the target domain, while the regular users do not.
In the target domain, members will have good results, but regular users will not.
In the source domain where all users have access, applying the prompt-based secret key will cause the accuracy for members to decrease by 5\%, which is undesired.
In contrast, applying the adapter-based secret key does not cause such issues and the accuracy for members will be almost the same (0.2\% improvement) as for the regular users.
}

\subsection{Ablation Study}\label{sec:ablation}
\begin{table}[t!]
\small
\setlength\tabcolsep{3pt}
\centering
\scalebox{0.9}{
\begin{tabular}{l c  c c }
\toprule
  & Source & Target & $\Delta$ \\
\midrule
UNTL & 77.4 & 35.3  & 42.1 \\
$\textrm{UNTL}_{\textrm{w/o Domain Clsf}}$ & 74.0 & 35.5  & 38.5\\
$\textrm{UNTL}_{\textrm{w/o MMD}}$ & 76.6 & 43.1  & 33.5\\

\bottomrule
\end{tabular}
}
\caption{Ablation studies of different distance methods in unsupervised non-transferable learning. $\Delta$ indicates the difference between the performance of the source and the target domain}
\label{tab:ablation}

\end{table}

\begin{table}[t!]
\small
\setlength\tabcolsep{3pt}
\centering
\scalebox{0.86}{
\begin{tabular}{l c c  c c }
\toprule
 & Source & Target & Target+Key & {$\Delta$}  \\
\midrule
PSK & 77.5  & 36.0  & 69.7 & 33.7\\

$\textrm{PSK}_{\textrm{w/o DC}}$& 76.9 &  39.5 & 69.5   & 30.0\\
$\textrm{PSK}_{\textrm{w/o MMD}}$& 70.3 & 41.6 & 40.8 & -0.8\\
\midrule
ASK & 77.7  & 36.1  & 74.4 & 38.3\\
$\textrm{ASK}_{\textrm{w/o DC}}$ & 70.7  & 64.4 & 66.2  & 1.8\\
$\textrm{ASK}_{\textrm{w/o MMD}}$ &  73.4  & 46.4 & 68.6 & 22.2\\

\bottomrule
\end{tabular}
}
\caption{Ablation studies for secret key-based UNTL. PSK and ASK denote prompt-based secret key and adapter-based secret key methods respectively. $\Delta$ indicates the difference between the performance between the Target+key and Target domains.}
\label{tab:key_ablation}
\end{table}

In this section, we investigate the impact of different UNTL losses: MMD and DC.
These two losses can maximize the discrepancy between representations of the source and target domains from different aspects.
The MMD loss will try to enlarge the average distance between the two domains, but the boundary may not be clear.
On the other hand, the DC loss makes up for the shortcomings of the MMD loss in terms of the boundary issue.
As Table~\ref{tab:ablation} shows, in UNTL, the difference between the performance of the source and target domain decreases when we remove either the MMD loss or the DC loss.
{\color{black} Based on this result, we use t-SNE \cite{JMLR:v9:vandermaaten08a} to visualize the representation distribution from the output of BERT of the source domain and the target domain. As Figure~\ref{dist} shows when only training with the DC or the MMD loss, the two distributions are close and the boundary can also be unclear. Only if we apply both losses, UNTL will be able to learn different distributions for the source and the target domains.}
Furthermore, from Table \ref{tab:key_ablation}, in the secret key-based methods, due to the similar initial representations of the target+key(prompt/adapter) and target domains, without using both losses to enlarge the distance, the model fails to perform well with the key (and badly without the key) in the target domain.
{\color{black}Besides, we also found that the prompt-based secret key method may rely more on the MMD loss, while the adapter-based secret key method tends to depend more on the DC loss.}
We speculate that the cause may be that: 1) in the prompt-based secret key method, the domain classifier can easily differentiate between the different domains based on the prompt pattern, but the representations will still be close in the continuous space without the MMD loss, whereas
2) in the adapter-based secret key method, the initial output embeddings of the adapter module are the same as the input ones.
Initially, the representations of the target+adapter domain and target domain could be highly similar to each other, resulting in a small MMD loss.
Thus, when only the MMD loss is used, the adapter may be stuck in the initial state and it is difficult to make progress on separating the two domains from each other during the fine-tuning phase.
{\color{black} On the other hand, the DC loss can offer stronger supervision than the MMD loss in terms of separating such two domains. Therefore, the DC loss could be playing a more significant role in the adapter-based secret key method.}


\section{Conclusion and Future Work}
In this paper, we present our UNTL method, which trains a model to maintain a good performance in the source domain whilst having degraded performance in the target domain.
Thereafter, we extend our approach with a secret key component that allows the restoration of the model performance in the target domain when the key is supplied, through two methods: prompt-based secret key and adapter-based secret key.
The experiments conducted on MultiNLI datasets suggest that our unsupervised non-transferable learning method can allow the model to perform differently in the source domain and the target domain.
The extensive experiments also demonstrate that our methods can effectively recover the ability of the model on the target domain with the specific secret key after non-transferable learning.

\textcolor{black}{For future works, we plan to extend our methods to incorporate multiple secret keys to achieve more than two user access levels by parameter-efficient methods~\cite{DBLP:conf/iclr/HeZMBN22} where we can first train our UNTL model, then freeze the parameters in the pretrained UNTL model, and train additional modules, such as prefix~\cite{DBLP:conf/acl/LiL20} and adapter~\cite{houlsby2019parameter}, to realize different user levels.
We also plan to explore other ways to degrade the performance in specific domains while maintaining good performance in other domains.}

\section*{Limitations}

In unsupervised non-transferable learning methods, after fine-tuning, the model tends to predict the same label for any input in the target domain.
In other words, when the model recognizes an input coming from the target domain, it tends to consistently assign a particular label.
We would like to highlight that, while our method is effective in the sense that it prevents the model from functioning well on the target domain, there is no guarantee that it would always yield ``worse performance'' as measured by accuracy as an evaluation metric.
Consider an extreme scenario where the labels in the target domain are highly unbalanced -- the domain consists of instances labeled with a particular label only.
At the same time, our model happens to predict that particular label for any input from the target domain.
In that case, the model may seemingly perform very well with a ``high accuracy''.
To resolve this known limitation, a different evaluation metric may be needed in order to properly assess the true performance of our model in target domains with unbalanced labels.

%


\section*{Ethics Statement}

Our work focuses on our unsupervised non-transferable learning in order to protect neural networks as intellectual property whilst making secure authorization more flexible with the secret key methods.
Nevertheless, we would like to point out that a malicious third-party neural network provider may utilize these methods for harmful purposes.
{\color{black} For example, the provider could use unsupervised non-transferable learning and secret key methods to insert an invisible backdoor into the model and extract private information from it.}

\section*{Acknowledgements}
\textcolor{black}{We would like to thank the anonymous reviewers, our meta-reviewer, and senior area chairs for their constructive comments and support on this work.
We would also like to thank Vanessa Tan for her help.
This research/project is supported by the National Research Foundation, Singapore under its AI Singapore Programme (AISG Award No: AISG-PhD/2021-08-007[T]).
}

\bibliography{anthology}
\bibliographystyle{acl_natbib}

\appendix

\section{Implementation Details}\label{sec:appendix}

\subsection{Network Architecture}
We use the {\em bert-base-uncased} model in Huggingface as our feature extractor.
As for the text classifier, domain classifier, and adapter module, the architecture of these modules is shown in Table~\ref{tab:archi}.

\begin{table}[ht!]
\small
\setlength\tabcolsep{3pt}
\centering
\scalebox{1.0}{
\begin{tabular}{l c  }
\toprule
 &   Architecture \\
\midrule
Text Classifier & Linear(768, 3) \\
\midrule
Domain Classifier & Linear(768, 2)\\
\midrule
 & Linear(64, 768) \\
Adapter & ReLU() \\
 & Linear(768, 64) \\

\bottomrule
\end{tabular}
}
\caption{The architecture of modules}
\label{tab:archi}
\end{table}

\subsection{Hyperparameters}
The learning rate of our experiment is shown in Table~\ref{tab:lr}.
We use three different seeds in our experiments: 20, 2022, and 2222.
The batch size is 256.
However, since the memory of the GPU is limited, we use Gradient Accumulation, a mechanism of PyTorch, to split a large batch of samples into several smaller batches of samples.
In our UNTL experiments, the gradient accumulation step is 2 and the small batch size is 128 so that the accumulated batch size is $128 * 2=256$.
We set the number of evaluation steps to 40.
We use 5 epochs in the baseline experiments while 8 epochs in our UNTL experiments.

\begin{table}[t!]
\small
\setlength\tabcolsep{3pt}
\centering
\scalebox{0.95}{
\begin{tabular}{l c  c c c c}
\toprule
 &   Bert & Text Cls & Domain Cls & Adapter \\
\midrule
Baseline & 5e-5 & 1e-3 & - & - \\
UNTL &  5e-5 & 15e-4 & 1e-3 & - \\
UNTL w/ prompt &5e-5 & 2e-3 & 1e-3 & -\\
 UNTL w/ adapter&  5e-5 & 2e-3 & 1e-3 & 1e-3\\

\bottomrule
\end{tabular}
}
\caption{Learning rates in different unsupervised non-transferable learning}
\label{tab:lr}
\end{table}
\begin{table}[t!]
\small
\setlength\tabcolsep{3pt}
\centering
\scalebox{1.05}{
\begin{tabular}{l c  c c c c}
\toprule
 &    $\alpha$ &$ \beta$ & $\lambda$& $c$ & $\omega$\\
\midrule

UNTL &              -  & 0.5 & 0.1 & 10.0   & 1.0\\
UNTL w/ prompt &    5.0 & 2.0 & 0.1 & 10.0 & 4.0\\
 UNTL w/  adapter & 10.0 & 1.5 & 0.1 & 10.0 & 2.0\\

\bottomrule
\end{tabular}
}
\caption{Hyperparameters in different unsupervised non-transferable learning}
\label{tab:hyper}
\end{table}

As for the hyperparameters, in our implementation, we apply a new scaling factor $\omega$ to control the cross entropy loss for text classification in order to balance the distance loss (especially in the secret key-based methods). The hyperparameters are shown in Table~\ref{tab:hyper}.

\subsection{Metric}
As the unsupervised non-transferable learning task aims to train the model to have a good performance in the source domain while performing badly in the target domain, we are more concerned with the difference between the performance of the source and the target domains. Therefore, we present a new metric based on it:
\begin{equation}
    \textrm{Difference}_{\textrm{UNTL}} = \textrm{Acc}_{\textrm{S}} - \textrm{Acc}_{\textrm{T}}
\end{equation}
where $\textrm{Acc}_{\textrm{S}}, \textrm{Acc}_{\textrm{T}}$ denote the accuracy in the source and the target domains respectively.

As for secret key-based UNTL, we aim to improve the performance of both the target+key(prompt/adapter) domain and the source domain while degrading the target domain performance. Therefore, we add a new difference between the performance of the key domain and the target domain to construct the metric as:
\begin{equation}
    \textrm{Difference}_{\textrm{Secret key}} = \textrm{Acc}_{\textrm{S}} + \textrm{Acc}_{\textrm{Key}} - 2 \textrm{Acc}_{\textrm{T}}
\end{equation}
where $\textrm{Acc}_{\textrm{Key}}$ denotes the accuracy in the key domain.
With such metrics, we select the best checkpoint based on the score over the development set.




\begin{table}[t!]
\normalsize
\setlength\tabcolsep{2pt}
\centering
\scalebox{0.8}{
\begin{tabular}{l ccccc}
\toprule
{Datasets} & \textsc{Book}  & \textsc{DvD} & \textsc{Elec} & \textsc{Kitchen}\\
\midrule
\#Train& 1,600 & 1,600 & 1,600 & 1,600\\
\#Valid& {\color{white}0,}200 & {\color{white}0,}200 & {\color{white}0,}200 & {\color{white}0,}200\\
\#Test& {\color{white}0,}200 & {\color{white}0,}200 & {\color{white}0,}200 & {\color{white}0,}200\\
\bottomrule
\end{tabular}
}
\caption{Statistics of polarity Amazon dataset}
\label{tab:polarityAmazonDataset}
\end{table}

\begin{table}[t!]
\normalsize
\setlength\tabcolsep{2pt}
\centering
\scalebox{0.8}{
\begin{tabular}{l ccccc}
\toprule
{Datasets} & \textsc{Beauty}  & \textsc{Book} & \textsc{Elec} & \textsc{Music}\\
\midrule
\#Train& 4,800 & 4,800 & 4,800 & 4,800\\
\#Valid& {\color{white}0,}600 & {\color{white}0,}600 & {\color{white}0,}600 & {\color{white}0,}600\\
\#Test& {\color{white}0,}600 & {\color{white}0,}600 & {\color{white}0,}600 & {\color{white}0,}600\\
\bottomrule
\end{tabular}
}
\caption{Statistics of ternary Amazon dataset}
\label{tab:terAmazonDataset}
\end{table}

\section{\textcolor{black}{Experiments on Additional Datasets}
}\label{sec:appexp}

\subsection{Datasets}
In this part, we present our experimental results on two additional datasets for sentiment analysis from \citet{DBLP:conf/acl/BlitzerDP07} (binary classification) and \citet{DBLP:conf/emnlp/HeLND18} (ternary classification).
We here denote them as polarity Amazon dataset and ternary Amazon dataset respectively. Table~\ref{tab:polarityAmazonDataset} and~\ref{tab:terAmazonDataset} show the statistics of them.

\begin{table}[t!]
\small
\setlength\tabcolsep{3pt}
\centering
\scalebox{1.0}{
\begin{tabular}{l c c c c}
\toprule
 Src\textbackslash Tgt & \textsc{Book}  & \textsc{DvD} & \textsc{Elec} & \textsc{Kitchen}\\
\midrule
 \multicolumn{1}{l}{\textsc{Book}} & 
$90.2_{ \pm 0.6 }$  &  $93.3_{ \pm 0.6 }$  &  $90.2_{ \pm 0.9 }$  &  $83.8_{ \pm 1.4 }$ \\ 
 \multicolumn{1}{l}{\textsc{DvD}} & 
 $89.5_{ \pm 1.8 }$  &  $92.2_{ \pm 1.2 }$  &  $89.7_{ \pm 1.2 }$  &  $87.0_{ \pm 0.8 }$ \\ 

 \multicolumn{1}{l}{\textsc{Elec}}& 
 $86.7_{ \pm 1.5 }$  &  $90.5_{ \pm 3.1 }$  &  $92.7_{ \pm 0.8 }$  &  $91.7_{ \pm 0.6 }$ \\ 
 \multicolumn{1}{l}{\textsc{Kitchen}} & 
 $84.3_{ \pm 2.1 }$  &  $92.2_{ \pm 1.0 }$  &  $93.8_{ \pm 0.9 }$  &  $91.0_{ \pm 0.8 }$ \\ 

\bottomrule
\end{tabular}
}
\caption{Performance of classification task in each domain when training only on the source domain of polarity Amazon dataset.}
\label{tab:sup_polarity}
\end{table}

\begin{table}[t!]
\small
\setlength\tabcolsep{3pt}
\centering
\scalebox{1.0}{
\begin{tabular}{l c c c c}
\toprule
 Src\textbackslash Tgt & \textsc{Book}  & \textsc{DvD} & \textsc{Elec} & \textsc{Kitchen}\\
\midrule
 \multicolumn{1}{l}{\textsc{Book}} & 
$89.2_{\pm 1.2}$ &  $50.2_{ \pm 0.2 }$  &  $50.2_{ \pm 0.2 }$  &  $50.0_{ \pm 0.0 }$ \\ 
 \multicolumn{1}{l}{\textsc{DvD}} & 
 $52.5_{ \pm 0.7 }$  & $92.3_{\pm 0.8}$ &  $50.7_{ \pm 0.5 }$  &  $50.0_{ \pm 0.0 }$ \\ 

 \multicolumn{1}{l}{\textsc{Elec}}& 
 $51.0_{ \pm 0.4 }$  &  $50.3_{ \pm 0.2 }$  & $93.9_{\pm 1.0}$ &  $52.0_{ \pm 0.4 }$ \\ 
 \multicolumn{1}{l}{\textsc{Kitchen}} & 
 $50.8_{ \pm 0.6 }$  &  $50.3_{ \pm 0.5 }$  &  $52.5_{ \pm 0.8 }$  & $90.7_{\pm 1.6}$\\
 
\bottomrule
\end{tabular}
}
\caption{Performance over the target domain for UNTL
on polarity Amazon dataset product
dataset.}
\label{tab:untl_polarity}
\end{table}

\begin{table}[t!]
\small
\setlength\tabcolsep{3pt}
\centering
\scalebox{1.0}{
\begin{tabular}{l c c c c}
\toprule
 Src\textbackslash Tgt & \textsc{Beauty}  & \textsc{Book} & \textsc{Elec} & \textsc{Music}\\
\midrule
 \multicolumn{1}{l}{\textsc{Beauty}} & 
$72.0_{ \pm 0.9 }$  &  $65.6_{ \pm 0.9 }$  &  $66.2_{ \pm 3.0 }$  &  $63.6_{ \pm 1.9 }$ \\ 
 \multicolumn{1}{l}{\textsc{Book}} & 

$63.7_{ \pm 2.3 }$  &  $78.7_{ \pm 1.1 }$  &  $55.0_{ \pm 1.8 }$  &  $62.3_{ \pm 1.6 }$ \\ 

 \multicolumn{1}{l}{\textsc{Elec}}& 
   $66.4_{ \pm 0.4 }$  &  $64.5_{ \pm 0.6 }$  &  $68.1_{ \pm 2.1 }$  &  $60.2_{ \pm 1.3 }$ \\  
 \multicolumn{1}{l}{\textsc{Music}} & 
 $59.2_{ \pm 2.2 }$  &  $68.4_{ \pm 0.3 }$  &  $58.0_{ \pm 2.3 }$  &  $69.8_{ \pm 1.6 }$ \\ 
 
\bottomrule
\end{tabular}
}
\caption{Performance of classification task in each domain when training only on the source domain of ternary Amazon dataset.}
\label{tab:sup_standard}
\end{table}

\begin{table}[t!]
\small
\setlength\tabcolsep{3pt}
\centering
\scalebox{1.0}{
\begin{tabular}{l c c c c}
\toprule
 Src\textbackslash Tgt & \textsc{Beauty}  & \textsc{Book} & \textsc{Elec} & \textsc{Music}\\
\midrule
 \multicolumn{1}{l}{\textsc{Beauty}} & 
 $71.4_{\pm 1.1}$ &  $33.6_{ \pm 0.4 }$  &  $34.9_{ \pm 1.2 }$  &  $33.6_{ \pm 0.1 }$ \\ 
 \multicolumn{1}{l}{\textsc{Book}} & 

 $33.4_{ \pm 0.1 }$  & $78.0_{\pm 0.8}$ &  $33.2_{ \pm 0.4 }$  &  $33.5_{ \pm 0.1 }$ \\ 

 \multicolumn{1}{l}{\textsc{Elec}}& 
  $33.1_{ \pm 1.3 }$  &  $33.3_{ \pm 0.1 }$  & $67.4_{\pm 2.5}$ &  $33.9_{ \pm 0.1 }$ \\ 
 \multicolumn{1}{l}{\textsc{Music}} & 
 $33.3_{ \pm 0.0 }$  &  $33.5_{ \pm 0.3 }$  &  $33.7_{ \pm 0.1 }$  & $69.2_{\pm 2.1}$\\ 
 
\bottomrule
\end{tabular}
}
\caption{Performance over the target domain for UNTL on ternary Amazon dataset.}
\label{tab:untl_standard}
\end{table}

\begin{table*}[t!]
\small
\setlength\tabcolsep{3pt}
\centering
\scalebox{0.92}{
\begin{tabular}{l c  c c c c }
\toprule
Source\textbackslash Target &  \textsc{Book}  & \textsc{DvD} & \textsc{Elec} & \textsc{Kitchen}\\
\midrule
\multicolumn{1}{c}{\textsc{Book}} &
$89.6_{ \pm 2.0 } \Rightarrow 88.8_{ \pm 1.9 }$  &  $51.3_{ \pm 0.8 } \Rightarrow 93.5_{ \pm 0.4 }$  &  $50.8_{ \pm 0.6 } \Rightarrow 90.2_{ \pm 0.8 }$  &  $50.0_{ \pm 0.0 } \Rightarrow 84.5_{ \pm 1.1 }$ \\ 
\multicolumn{1}{c}{\textsc{DvD}} & 
 $52.0_{ \pm 0.4 } \Rightarrow 90.8_{ \pm 1.4 }$  &  $93.7_{ \pm 1.0 } \Rightarrow 92.6_{ \pm 1.5 }$  &  $50.5_{ \pm 0.0 } \Rightarrow 88.0_{ \pm 3.2 }$  &  $50.2_{ \pm 0.6 } \Rightarrow 88.5_{ \pm 0.0 }$ \\ 
 
\multicolumn{1}{c}{\textsc{Elec}} &
$51.7_{ \pm 0.6 } \Rightarrow 86.5_{ \pm 1.8 }$  &  $50.5_{ \pm 0.4 } \Rightarrow 90.5_{ \pm 1.1 }$  &  $93.6_{ \pm 0.9 } \Rightarrow 91.9_{ \pm 1.3 }$  &  $53.3_{ \pm 0.8 } \Rightarrow 91.2_{ \pm 1.2 }$ \\ 
\multicolumn{1}{c}{\textsc{Kitchen}} & 
 $51.2_{ \pm 0.6 } \Rightarrow 89.2_{ \pm 3.3 }$  &  $50.8_{ \pm 0.5 } \Rightarrow 93.0_{ \pm 1.2 }$  &  $52.2_{ \pm 1.8 } \Rightarrow 94.0_{ \pm 0.7 }$  &  $90.9_{ \pm 1.0 } \Rightarrow 88.7_{ \pm 2.1 }$ \\
\bottomrule
\end{tabular}
}
\caption{Performance of prompt-based secret key UNTL model on polarity Amazon dataset. The left of the right arrow shows the accuracy (\%) of the model when the prompt-based key was not attached to the input, and the right is the precision with the key.}
\label{tab:polarity_prompt}
\end{table*}


\begin{table*}[t!]
\small
\setlength\tabcolsep{3pt}
\centering
\scalebox{0.92}{
\begin{tabular}{l c  c c c c }
\toprule
Source\textbackslash Target &  \textsc{Beauty}  & \textsc{Book} & \textsc{Elec} & \textsc{Music}\\
\midrule
\multicolumn{1}{c}{\textsc{Beauty}} &
$71.5_{ \pm 1.7 } \Rightarrow 67.0_{ \pm 3.3 }$  &  $33.4_{ \pm 0.2 } \Rightarrow 67.4_{ \pm 1.9 }$  &  $34.9_{ \pm 0.4 } \Rightarrow 64.6_{ \pm 0.9 }$  &  $33.8_{ \pm 0.3 } \Rightarrow 65.3_{ \pm 3.7 }$ \\ 
\multicolumn{1}{c}{\textsc{Book}} & 
$33.3_{ \pm 0.0 } \Rightarrow 64.7_{ \pm 2.8 }$  &  $78.6_{ \pm 0.7 } \Rightarrow 73.9_{ \pm 2.9 }$  &  $33.1_{ \pm 0.3 } \Rightarrow 60.3_{ \pm 1.2 }$  &  $34.3_{ \pm 0.1 } \Rightarrow 57.1_{ \pm 2.9 }$ \\ 
 
\multicolumn{1}{c}{\textsc{Elec}} &
$34.3_{ \pm 0.2 } \Rightarrow 67.1_{ \pm 1.3 }$  &  $33.2_{ \pm 0.4 } \Rightarrow 69.3_{ \pm 0.7 }$  &  $66.7_{ \pm 1.2 } \Rightarrow 64.1_{ \pm 3.8 }$  &  $33.9_{ \pm 0.3 } \Rightarrow 60.4_{ \pm 2.3 }$ \\ 
\multicolumn{1}{c}{\textsc{Music}} & 
$33.5_{ \pm 0.2 } \Rightarrow 62.7_{ \pm 0.8 }$  &  $33.4_{ \pm 0.5 } \Rightarrow 68.2_{ \pm 0.7 }$  &  $33.7_{ \pm 0.1 } \Rightarrow 64.0_{ \pm 1.7 }$  &  $66.6_{ \pm 1.6 } \Rightarrow 66.7_{ \pm 1.6 }$ \\ 
\bottomrule
\end{tabular}
}
\caption{Performance of prompt-based secret key UNTL model on ternary Amazon dataset. The left of the right arrow shows the accuracy (\%) of the model when prompt-based key was not attached to the input, and the right is the precision with the key.}
\label{tab:standrad_prompt}
\end{table*}


\begin{table*}[t!]
\small
\setlength\tabcolsep{3pt}
\centering
\scalebox{0.92}{
\begin{tabular}{l c  c c c}
\toprule
Source\textbackslash Target & \textsc{Book}  & \textsc{DvD} & \textsc{Elec} & \textsc{Kitchen}\\
\midrule
\multicolumn{1}{c}{\textsc{Book}} & 
$90.3_{ \pm 0.9 } \Rightarrow 91.8_{ \pm 1.4 }$  &  $51.7_{ \pm 1.0 } \Rightarrow 94.0_{ \pm 0.0 }$  &  $50.8_{ \pm 0.2 } \Rightarrow 89.0_{ \pm 0.4 }$  &  $50.5_{ \pm 0.7 } \Rightarrow 87.2_{ \pm 1.4 }$ \\ 

\multicolumn{1}{c}{\textsc{DvD}} & 
 $52.7_{ \pm 0.6 } \Rightarrow 91.2_{ \pm 1.3 }$  &  $92.7_{ \pm 1.5 } \Rightarrow 93.4_{ \pm 1.6 }$  &  $50.5_{ \pm 0.4 } \Rightarrow 90.3_{ \pm 0.5 }$  &  $50.0_{ \pm 0.0 } \Rightarrow 88.3_{ \pm 1.6 }$ \\ 
 
\multicolumn{1}{c}{\textsc{Elec}} & 
 $50.5_{ \pm 2.7 } \Rightarrow 88.7_{ \pm 1.2 }$  &  $51.2_{ \pm 0.6 } \Rightarrow 90.8_{ \pm 0.9 }$  &  $93.6_{ \pm 1.3 } \Rightarrow 93.7_{ \pm 0.9 }$  &  $53.2_{ \pm 0.6 } \Rightarrow 91.3_{ \pm 0.6 }$ \\ 

\multicolumn{1}{c}{\textsc{Kitchen}} & 
 $51.3_{ \pm 0.6 } \Rightarrow 89.5_{ \pm 2.9 }$  &  $51.0_{ \pm 0.4 } \Rightarrow 92.3_{ \pm 0.6 }$  &  $53.7_{ \pm 0.8 } \Rightarrow 93.0_{ \pm 0.7 }$  &  $90.9_{ \pm 1.4 } \Rightarrow 91.2_{ \pm 1.3 }$ \\ 

\bottomrule
\end{tabular}
}
\caption{Performance of adapter-based secret key UNTL model on polarity Amazon-product-review dataset. The left of the right arrow shows the accuracy (\%) of the model when the input adapter was not applied, and the right is the precision with the adapter.}
\label{tab:polarity_adapter}
\end{table*}

\begin{table*}[t!]
\small
\setlength\tabcolsep{3pt}
\centering
\scalebox{0.92}{
\begin{tabular}{l c  c c c}
\toprule
Source\textbackslash Target & \textsc{Beauty}  & \textsc{Book} & \textsc{Elec} & \textsc{Music}\\
\midrule
\multicolumn{1}{c}{\textsc{Beauty}} & 
$72.2_{ \pm 1.1 } \Rightarrow 72.1_{ \pm 1.2 }$  &  $33.3_{ \pm 0.2 } \Rightarrow 70.1_{ \pm 0.4 }$  &  $34.9_{ \pm 0.3 } \Rightarrow 68.3_{ \pm 1.0 }$  &  $33.6_{ \pm 0.1 } \Rightarrow 66.1_{ \pm 1.5 }$ \\ 

\multicolumn{1}{c}{\textsc{Book}} & 
$33.4_{ \pm 0.1 } \Rightarrow 68.6_{ \pm 0.7 }$  &  $77.5_{ \pm 0.8 } \Rightarrow 77.6_{ \pm 0.8 }$  &  $33.1_{ \pm 0.4 } \Rightarrow 63.5_{ \pm 1.0 }$  &  $33.8_{ \pm 0.3 } \Rightarrow 67.1_{ \pm 2.4 }$ \\ 
\multicolumn{1}{c}{\textsc{Elec}} & 
$34.4_{ \pm 0.5 } \Rightarrow 69.4_{ \pm 0.8 }$  &  $32.4_{ \pm 1.6 } \Rightarrow 71.4_{ \pm 1.8 }$  &  $67.9_{ \pm 1.5 } \Rightarrow 68.1_{ \pm 1.6 }$  &  $33.8_{ \pm 0.3 } \Rightarrow 64.3_{ \pm 1.6 }$ \\ 

\multicolumn{1}{c}{\textsc{Music}} & 
$33.4_{ \pm 0.1 } \Rightarrow 62.3_{ \pm 1.7 }$  &  $33.8_{ \pm 0.0 } \Rightarrow 74.1_{ \pm 2.7 }$  &  $33.4_{ \pm 0.2 } \Rightarrow 63.7_{ \pm 0.8 }$  &  $69.1_{ \pm 1.8 } \Rightarrow 69.3_{ \pm 1.5 }$ \\

\bottomrule
\end{tabular}
}
\caption{Performance of adapter-based secret key UNTL model on standard Amazon-product-review dataset. The left of the right arrow shows the accuracy (\%) of the model when the input adapter was not applied, and the right is the precision with the adapter.}
\label{tab:standrad_adapter}
\end{table*}

\subsection{Performance for UNTL}
Table~\ref{tab:sup_polarity} and \ref{tab:untl_polarity} show the performance for baseline results and the results for UNTL in the polarity Amazon dataset. Similarly, Table~\ref{tab:sup_standard} and \ref{tab:untl_standard} show the results in the ternary Amazon dataset. From the results above, we can find our UNTL method can maintain similar performance as the baseline and degrade the performance in the target domain to nearly random choice (33.3\% in the ternary classification tasks and 50.0\% in the polarity classification tasks).

\subsection{Performance for UNTL with Secret Keys}
Table~\ref{tab:polarity_prompt} and \ref{tab:standrad_prompt} show the performance for UNTL with Prompt-based Secret Key in the polarity Amazon dataset and the ternary Amazon dataset respectively. 
As for the Adapter-based Secret Key, Table~\ref{tab:polarity_adapter} and \ref{tab:standrad_adapter} show the performance in these two datasets respectively. After applying the secret keys, we can see that the performance in the target domain can be restored to better results. Comparing two different kinds of secret keys, we find that the adapter-based models can get better results in the target domain than the prompt-based models.


 
 

\end{document}